\documentclass[10pt,twocolumn,letterpaper]{article}

\usepackage{cvpr}              

\usepackage[dvipsnames]{xcolor}

\definecolor{cvprblue}{rgb}{0.21,0.49,0.74}
\usepackage[pagebackref,breaklinks,colorlinks,citecolor=cvprblue]{hyperref}

\def\confName{CVPR}

\title{Solution for CVPR 2024 UG2+ Challenge Track on All Weather Semantic Segmentation}

\author{Jun Yu, Yunxiang Zhang, Fengzhao Sun, Leilei Wang, Renjie Lu\\
University of Science and Technolog of China\\
{{\tt\small harryjun@ustc.edu.cn,} {\tt\small \{mesa,sunfz,sa22218164,renjielu\}@mail.ustc.edu.cn}}
}

\begin{document}
\maketitle
\begin{abstract}
In this report, we present our solution for the semantic segmentation in adverse weather, in UG2+ Challenge at CVPR 2024. To achieve robust and accurate segmentation results across various weather conditions, we initialize the InternImage-H backbone with pre-trained weights from the large-scale joint dataset and enhance it with the state-of-the-art Upernet segmentation method. Specifically, we utilize offline and online data augmentation approaches to extend the train set, which helps us to further improve the performance of the segmenter. As a result, our proposed solution demonstrates advanced performance on the test set and achieves 3rd position in this challenge.

\end{abstract}    
\begin{figure}[t]
  \centering
  \includegraphics[width=1\linewidth]{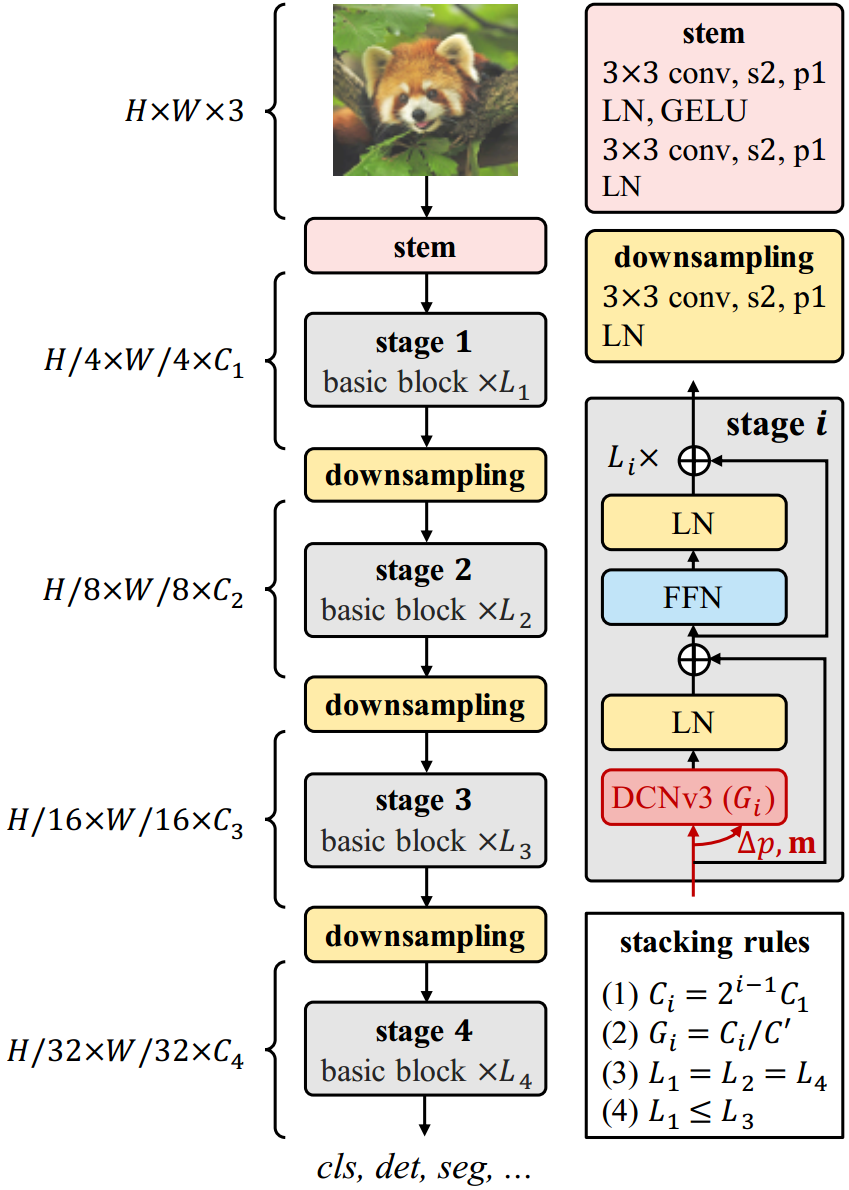} 
  \caption{ Overall Architecture of InternImage, where the core operator is DCNv3, and the basic block composes of layer normalization (LN) \cite{ba2016layer} and feed-forward network (FFN) \cite{vaswani2017attention} as transformers, the stem and downsampling layers follows conventional CNN’s designs}
  \label{fig:1}
\end{figure}

\section{Introduction}
Semantic segmentation boasts a rich history of applications in autonomous driving\cite{Ess2009SegmentationBasedUT,Nekrasov2018RealTimeJS,Siam2018RTSegRS}, robotics\cite{Kim2018IndoorSS,Milioto2018BonnetAO,Milioto2017RealTimeSS}, and scene understanding\cite{Cordts2016TheCD,Li2019IndoorSU}. Although contemporary methods have demonstrated remarkable efficacy on standard benchmarks, i.e. ADE20K\cite{Zhou2016SemanticUO}, Cityscapes\cite{Cordts2016TheCD}, real-world environments often present complexities and challenges, such as adverse weather conditions. When confronted with images exhibiting visual degradations, specifically those captured under such unfavorable conditions, their performance correspondingly deteriorates. 

To address these issues, researchers\cite{Kerim2022SemanticSU,Sakaridis2017SemanticFS} have provided datasets and methods aimed at studying the effects of natural phenomena. However, due to the difficulty of capturing paired datasets in controlled settings, existing datasets often rely on synthetic weather effects or include misalignments between degraded and clear-weather images in the underlying scene. Futher more, \cite{Gella2023WeatherProofAP} introduces the WeatherProof Dataset, a semantic segmentation dataset specifically designed for weather-degraded scenes. Training on this paired dataset can significantly improve model performance in adverse weather conditions.

The purpose of this challenge is to investigate weather phenomena in real-world scenarios and to inspire innovative advancements in semantic segmentation methods for images affected by weather conditions.

In this work, we address real-world complex scenarios by making targeted improvements in data processing, algorithm design, and model optimization, achieving state-of-the-art segmentation performance. Our contributions are outlined below:

\begin{itemize}
\item \textbf{Data Augmentation}: We utilize offline and online data augmentation approaches to extend the training set, which helps to further improve the performance of the segmenter.
\item \textbf{Algorithm Fine-tuning}: We employ large, foundational model Internimage as our segmenter
with UperNet as the backbone network, which has led to drastically improved performance in the task of semantic segmentation.
\item \textbf{Model Optimization}: We implement advanced model fusion strategies, specifically by combining the results of multiple models through techniques such as voting, to refine model accuracy and robustness in weather-degraded scenes.
\end{itemize}

\section{Method}

In this section,we introduce each component that we have attempted in our approach. 

\subsection{Encoder}
As shown in Figure \ref{fig:1}, Internimage \cite{wang2023internimage} is a new large-scale NN-based foundation model which effectively scales to over 1 billion parameters and 400 million training images and achieves comparable or even better performance than state-of-the-art ViTs. Different from the recent CNNs that focus on large dense kernels, InternImage takes deformable convolution as the core operator, so that it not only has the large effective receptive field required for downstream tasks such as detection and segmentation, but also has the adaptive spatial aggregation conditioned by input and task information. As a result, InternImage reduces the strict inductive bias of traditional CNNs and makes it possible to learn stronger and more robust patterns with large-scale parameters from massive data like ViTs. With its powerful object representation capabilities, Internimgage has demonstrated impressive performance on various representative computer vision tasks. For example, InternImage-H has achieved an improvement of 89.6\% top-1 accuracy on ImageNet , and achieves 62.9\% mIoU and 65.4\% mAP on the challenging downstream benchmarks ADE20K  \cite{Zhou2016SemanticUO} and COCO \cite{lin2014microsoft} , respectively. Besides, it’s worth mentioning that internimgage-H is one of the few open-source large models available.  As a result, we have to implement it ourselves .

\subsection{Decoder}

UPerNet is a widely adopted network architecture for semantic segmentation, which integrates the ideas of Pyramid Pooling Module (PPM) \cite{zhao2017pyramid} and Feature Pyramid Network (FPN) \cite{lin2017feature}. It effectively fuses feature information from different scales, enhancing the segmentation capability of models for objects at various scales. It has achieved outstanding performance on multiple segmentation benchmarks.Here, InternImage has been enhanced for UPerNet by incorporating layer normalization (LN) \cite{ba2016layer} and feed-forward network (FFN) \cite{vaswani2017attention} , and utilizing GELU \cite{hendrycks2016gaussian} as the activation function.

\section{Experiments}

\subsection{Dataset}

The competition's dataset: WeatherProof dataset– a semantic segmentation dataset
with over 174.0K images. It is the first with high quality semantic segmentation labels, with accurately aligned and paired clear and adverse weather images for more accurate evaluation under controlled settings. The dataset includes 513 sets of images, each consisting of 300 pairs of clean and adverse condition images. Additionally, we used 38 sets of images for the validation phase.

\subsection{Data Augmentation}
Due to the limitations of the train set and the adverse conditions, we employ various data augmentation techniques,including both offline and online approaches. 
The offline data augmentation primarily focuses on modifying the contrast and brightness of the images, which effectively expands the train set. During the training process, we apply different online data augmentation methods for better results. We employ RandomCrop, RandomFlip and padding as the data augmentation methods. \subsection{Implementation Details}
The training process is initialized by the
InternImage-H model pretrained on ADE20K \cite{Zhou2016SemanticUO}. We train our model
with 8 NVIDIA Tesla V100 GPUs for 6000 iterations, using AdamW\cite{loshchilov2017decoupled} as optimizer with  an initial learning rate of 0.00002, weight decay of 0.05 and with a crop size of 960.

\subsection{Final Results}
\begin{figure*}[t]
  \centering
  \includegraphics[width=1\linewidth]{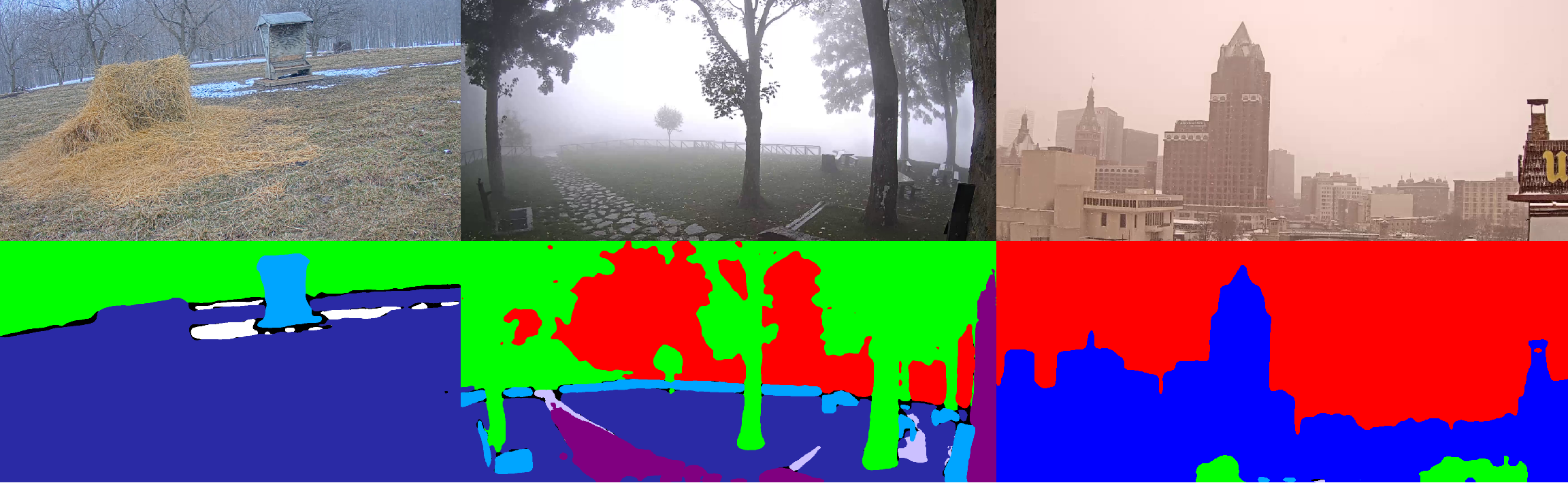} 
  \caption{Visualization of Select Segmentation Outputs}
  \label{fig:2}
\end{figure*}

Our optimal model achieved superior performance on the validation set after 3500 iterations employing InternImage. Subsequently, we conducted hard voting on the output results of this optimal model along with some closest models. The aggregated output achieved an mIoU of 0.4371 on the test set. Detailed metrics for our ensemble model are presented in Table \ref{tab:commands}, showcasing a significant enhancement in segmentation efficacy through the utilization of model fusion techniques, such as voting. Furthermore, visualizations of select segmentation outputs are depicted in Figure \ref{fig:2}

\begin{table}
  \centering
  \caption{Comparison Results of Samples Used for Fusion and Hard Voting Outcomes}
  \label{tab:commands}
  \begin{tabular}{ccl}
    \toprule
     models& test mIoU\\
    \midrule
    \text{iter 3000}& 0.4040\\
    \text{iter 3500}&  0.4198\\
    \text{iter 4000}& 0.4184 \\
    \text{Voting results}& 0.4371\\
    \bottomrule
  \end{tabular}
\end{table}

\section{Conclusion}

In this report, we propose a InternImage-based segmentation method for the semantic segmentation in adverse weather. To cope with the limitations of the dataset and adverse conditions, we employ offline and online data augmentation approaches to extend the train set, and utilize InternImage as our semantic segmenter with UperNet as the backbone network. The results show that our proposed solution achieves outstanding performance on the test set in UG2+ Challenge 2024.
{
    \small
    \bibliographystyle{ieeenat_fullname}
    \bibliography{main}
}


\end{document}